\documentclass[nohyperref]{article}

\usepackage{microtype}
\usepackage{graphicx}
\usepackage{subfigure}
\usepackage{booktabs}
\usepackage{enumitem}

\usepackage{hyperref}

\usepackage[accepted]{icml2022}

\usepackage{amsmath}
\usepackage{amssymb}
\usepackage{mathtools}
\usepackage{commands}
\usepackage[accepted]{icml2022}

\icmltitlerunning{Explainable AI for survival analysis: a median-SHAP approach}

\begin{document}

\twocolumn[
\icmltitle{Explainable AI for survival analysis: a median-SHAP approach}

\begin{icmlauthorlist}
\icmlauthor{Lucile Ter-Minassian}{yyy}
\icmlauthor{Sahra Ghalebikesabi}{yyy}
\icmlauthor{Karla Diaz-Ordaz}{comp,sch}
\icmlauthor{Chris Holmes}{yyy,sch}
\end{icmlauthorlist}

\icmlaffiliation{yyy}{University of Oxford}
\icmlaffiliation{comp}{The London School of Hygiene \& Tropical Medicine}
\icmlaffiliation{sch}{The Alan Turing Institute}

\icmlcorrespondingauthor{Lucile Ter-Minassian}{lucile.ter-minassian@spc.ox.ac.uk}

\icmlkeywords{Interpretable ML, Healthcare}

\vskip 0.3in
]
\printAffiliationsAndNotice{\icmlEqualContribution} 

\begin{abstract}
With the adoption of machine learning into routine clinical practice comes the need for Explainable AI methods tailored to medical applications. Shapley values have sparked wide interest for locally explaining models. Here, we demonstrate their interpretation strongly depends on both the summary statistic and the estimator for it, which in turn define what we identify as an 'anchor point'. We show that the convention of using a mean anchor point may generate misleading interpretations for survival analysis and introduce median-SHAP, a method for explaining black-box models predicting individual survival times. 
\end{abstract}

\section{Introduction}

Recent years have seen a rapid growth in the Explainable AI (XAI) literature. Given the great variety of explanation models and the diversity of interpretations they can generate \cite{sundararajan2020many}, decision-makers are increasingly demanding that XAI methods be designed for a single application. Contrasting with one-size-fits all solutions, such tailored XAI techniques would better fulfill the precise requirements of the task they're meant for \cite{arrieta2020explainable}. Here, we introduce a specific method for explaining black-box survival analysis models, which is based on Shapley values.

Shapley values (SV) have become a gold standard for local model explainability. They compute feature attributions by quantifying the change in model output when dropping certain feature values and sampling them from a reference distribution. The choice of a reference probability distribution on feature values has been the subject of multiple debates \cite{aas2019explaining, janzing2020feature, frye2020shapley}. In contrast, we argue that a key challenge of SV arises from the convention of both using (i) the expectation as a summary statistic for the distribution of model outcomes and (ii) the sample mean to estimate it, and that this question has been overlooked. Most importantly, we show these parameters jointly induce a mean 'anchor point' against which comparisons are made, and how such an anchor point isn't adapted the time-to-event models. Ultimately, we demonstrate that using the original formulation of Shapley values can generate misleading interpretations when the distribution of the value function is right-skewed, which is the case for survival analysis models as some individuals may not be experiencing the event throughout the study \cite{rao2007survival, ying1995survival}. 

To resolve this problem, we introduce median-SHAP, which uses the median as the summary statistic and for estimation. We demonstrate the benefits of this approach for explaining survival analysis models that output predicted survival times from a set of features. Note that such models differ from inferential methods (e.g. the Cox proportional hazards model \cite{lin1989robust}) for understanding the drivers of survival. For point prediction models, the median predicted survival time is more commonly reported than the mean. 

To the best of our knowledge, our method is the first additive feature attribution method tailored to survival analysis. In contrast, SurvLIME \cite{utkin2020survlime, kovalev2020survlime} is a local linear approximation based on the Cox proportional hazards model.

\paragraph{Contributions} The gaps in the existing literature motivate the following contributions of this paper: 
\vspace*{-0.2cm}
\begin{enumerate}[leftmargin=4mm, itemsep=0.5mm]
    \item We shed a new light on the discussion around SV by interrogating the notions of summary statistics and mean estimators within Shapley values and the subsequent 'anchor points' they induce. We identify the anchor point as the instance our observation of interest is compared against, and explain why it is essential for the interpretation of SV.
    \item We show how using a mean anchor point can generate misleading explanations of survival analysis models
    \item We introduce median-SHAP, an explanation model specific to survival analysis. Our method is based on observational SV and uses the median as a summary statistic. We experimentally show that our method has improved interpretability and robustness compared with the original SV.
\end{enumerate}


\section{Shapley Values} 
\label{sec:shapley_values}

We consider a model $\blackbox:\mathbb{R}^{\numfeats}\rightarrow\mathbb{R}^{\numout}$ and aim to explain the prediction of an instance $\localobs\in\mathbb{R}^{\numfeats}$, given only black-box access to the model. Shapley values quantify the contribution of input features $\{1,\ldots,\numfeats\}$ to the prediction of a complex model $\blackbox:\mathbb{R}^{\numfeats}\rightarrow\mathbb{R}^{\numout}$ at an instance $x$ as follows: 
\begin{align}\label{eq:efficiency}
\blackbox(x)=\phi^{\blackbox}_{0} +\sum_{i=1}^{M} \phi^{\blackbox}_{i}(x)
\end{align}
where $\phi^{\blackbox}_{j}(x)$ is the Shapley value of feature $j$ to $\blackbox(x)$ and $\phi^{\blackbox}_{0}=\mathbb{E} [f(X)]$ is the prediction averaged over the observed data distribution. The attribution of a feature $j$ is computed from the difference in value function $v_f$ comparing when the feature $j$ is equal to the value $x_j$ with when it is removed from the coalition. We denote this difference $\Delta v_f(S, j, x) = v_f(S \cup \{j\}, x) - v_f(S, x)$. When a feature is included in the coalition its value is set to the observed instance value $\localobs_{\includedfeats}$. When a feature is not in the coalition its value is sampled from a reference distribution $\refdistvar{}{\includedfeats}$. \newline
\centerline{$v_f(\feat{\includedfeats})=\expect_{\refdistvar{}{\includedfeats}}[\blackbox({\localobs_{\includedfeats}, \varimputed_{\droppedfeats}})]$} \newline
for $\droppedfeats:=\{1,\ldots,\numfeats\} \backslash \includedfeats$ with $({\localobs_{\includedfeats}, \localobs_{\droppedfeats}})$ denoting the concatenation of its two arguments. 
To account for the dependence with other features, one takes the difference in value function $v$ averaged over all possible coalitions $\includedfeats$ of features excluding feature $j$.  Ultimately, a binomial weight $\frac{{|\includedfeats|!(\numfeats-|\includedfeats|-1)!}}{m!}$ is added to recover the original Shapley values which account for all possible orderings. All in all, the Shapley values of feature $j$ is defined as follows:
\begin{align*}
    \phi_j^\blackbox(x) 
    &=   \frac{1}{\numfeats} \sum_{\substack{|\includedfeats|\in \\ \{0,\ldots,\numfeats-1\}}} \left[\frac{1}{\binom{\numfeats-1}{|\includedfeats|}} \sum_{\substack{\includedfeats\subseteq \{1,\ldots,\numfeats\}/j\; \\\text{with }|S|=j}} \Delta v_f(S, j, x)\right] \\
    &= \expect_{S}\left[\expect_{\refdistvar{}{\includedfeats}}[\blackbox({\localobs_{\includedfeats}, \varimputed_{\droppedfeats}})]\right]
\end{align*}
Further, note that the two expectations can be taken in whichever order. We refer to the \emph{expectation as a summary statistic} to describe the fact that the distribution of model outcomes (whether over coalitions or reference points) is summarised using the central location. 

The choice of a reference distribution for SV has been subject to recent debates \cite{aas2019explaining, janzing2020feature, merrick2020explanation}. Interventional SV \cite{lundberg2017unified, janzing2020feature} define $\refdistvar{}{\includedfeats}:=\dist({\varimputed})$ where $\dist$ denotes the marginal data distribution. Observational Shapley values \cite{aas2019explaining} set the reference distribution equal to the conditional distribution given $\localobs_{\includedfeats}$ $\refdistvar{}{\includedfeats}:=\dist({\varimputed|\varimputed_{\includedfeats}=\localobs_{\includedfeats}})$. Observational SV are often described as 'true to the data' because they do not break the correlations between features. Moreover, this makes them robust to adversarial attacks, as their computation does not involve evaluations on out-of-distribution instances i.e.\ concatenated inputs which were built by sampling from a marginal distribution.

At this stage, it is important to note that, in reality, SV assess the importance of a feature \emph{value}, and not a feature. In other words, the central question of SV is 'By how much did the feature $j$ shift the local model's prediction $f(x)$ away (in any direction) from the average prediction $\phi_0^f$, once feature $j$ was set to $x_j$?'. By implementing Equation \ref{eq:efficiency} -- also known as the \textit{Efficiency} property -- we are able to derive bespoken drift from the SV, e.g. 'mode shift'. Consequently, as feature attributions result from comparisons between $f(x)$ and $\phi_0^f$, our instance $x$ is being compared with a synthetic input for which the model prediction is equal to $\phi^f_0(x)$. We call this synthetic input the \emph{anchor point} in the following. Such comparisons are not readily interpretable as the anchor point is not an actual realisation of the population. 

Finally, given that Shapley values are Monte Carlo approximated in practice, the expectation over the reference distribution is estimated by the mean over a finite set of $\numimp$ observations $\{\imputed_{i, \droppedfeats}\}_{i=1}^{\numimp}$ from the reference distribution, henceforth called 'reference points'  $\refdistvar{\includedfeats}{\droppedfeats}$,
\begin{align*}
    \widehat{\phi}(j) &=  \frac{1}{\numfeats} \sum_{\includedfeats\subseteq \{1,\ldots,\numfeats\}/j}
    [\frac{1}{\binom{\numfeats-1}{|\includedfeats|}}  (\frac{1}{\numimp}\sum_{i=1}^{\numimp}\blackbox(\localobs_{\includedfeats \cup j}, \imputed_{i, \droppedfeats/j}) \\ 
    &- \frac{1}{\numimp}\sum_{i=1}^{\numimp}\blackbox(\localobs_{\includedfeats}, \imputed_{i, \droppedfeats \cup j})].
\end{align*}
We will refer to this as using the \emph{sample mean as an estimator} of the expectation, or central location. 

\section{Summary statistics, estimators, and anchor points}
\label{sec:anchor}

In the following, we differentiate between the drawbacks of the expectation as a summary statistic, the sample mean over reference points as its estimator and the previously defined an anchor point.

\subsection{Drawbacks of the expectation as a summary statistic}

When averaging the difference in value function $\Delta v_f$ over all coalitions, the original Shapley formulation summarises the distribution of the model outcomes of concatenated inputs by taking their expectation. First, taking the expectation to summarise central location is problematic when the distribution is skewed. Second, the value of the expectation -- and not its relative position within the sample distribution -- can lead to misleading interpretations. For instance, if the single reference values are centred around zero, this may give the impression that the corresponding feature has a negligible impact on the model outcome. For instance, consider a model \newline
\centerline{$\blackbox({\localobs})=1000\localobs_1+\localobs_2$} \newline
at observation $x=(0,1)$ and where marginally $x_1\sim\Normal(0, 1)$ and $x_2\sim\Normal(0, 1)$. 
Adding $x_1$ to the empty coalition does not, on average, make a difference in terms of model outcome as on the one hand we marginalise over $x_1$ which is centered on 0, and on the other hand we specify $x_1=0$. Similarly, model outcomes are be equal in expectation for the coalitions $\{x_2\}$ and $\{x_1,x_2\}$. \cite{fryer2021shapley} argue this is why Shapley values should not be used for feature selection.  We then have Shapley values: \newline 
\centerline{$\phi^{f}_{X_1}(x_1,x_2)=0$ \ \ \ \ \ \ \ \ \ \ \ \ $\phi^f_{X_2}(x_1,x_2)=1$}\newline 
which can be misleading. 
Challenges can also arise from taking the expectation over feature coalitions, as shown by Merrick et.al \yrcite{merrick2020explanation}. To solve this problem, the authors propose to cluster the single reference Shapley values and return the cluster means. Nonetheless, multidimensional clustering algorithms used on one-dimensional data have several shortcomings, including the fact that two clusters might be of considerably different size. Instead we remark that since we are interested in summarising a distribution, we should thus use summary statistics that are robust to distribution shifts, such as quantiles.

\subsection{Drawbacks of the mean estimator}

Using the sample mean over reference points as an estimator of central location is challenging when in the presence of heavy-tailed data distributions and outliers. Consider, for instance a black-box model \newline
\centerline{$\blackbox(\localobs)=\localobs+\epsilon$} 
where $\epsilon$ is Cauchy noise. Here, the Shapley value at $\localobs$ should be $\infty$ in theory. In practical applications, an adversary might return extreme values for unlikely observations, i.e. $\blackbox(\localobs)=-\localobs-1000\cdot\indicate{\localobs > 0.99}, \localobs\sim\Uniform[0, 1]$. The extreme values with small probability push the mean change in model outcome of including the feature to the positive at any observation with $\localobs$ outside of $[0.99, 1]$. Such an effect is all the more likely to happen for marginal Shapley values, as the model estimator can suffer under high variance. 

\subsection{Drawbacks of a global mean 'anchor point'}

Taking the expectation as a summary statistic and the sample as its estimator together imply that comparisons are made \wrt the global mean, which acts as an anchor point. This has negative impacts on the interpretability of the resulting attributions as the average model outcome over reference points does not link to any single particular individual. Put differently, in a clinical example we would compute the respective SV for a patient comparing them with an 'artificial' patient whose model outcome would be equal to the global mean $\phi_0$. Further, when this distribution is marginal, the anchor point might not relate to a plausible instance. This poses a problem for clinical management, as patients may want to understand the influence of modifiable features (e.g. smoking, physical activity) to lengthen their survival.


\section{median-SHAP for predictions of median survival times}
\label{sec:median-SHAP}

Our proposed method is a local explanation model based on \textit{observational} Shapley values tailored to black-box models that predict a median survival time. median-SHAP takes the median of the expected change in model outcome for a coalition $\includedfeats$, that is
\begin{align*}
    & \widehat{\phi}(j)^{med} = \frac{1}{\numfeats} \sum_{\includedfeats\subseteq \{1,\ldots,\numfeats\}/j} [\frac{1}{\binom{\numfeats-1}{|\includedfeats|}} \\ & (med(\{\blackbox(\localobs_{\includedfeats \cup j}, \imputed_{i, \droppedfeats/j})\}_{i=1}^{\numimp})- med(\{\blackbox(\localobs_{\includedfeats}, \imputed_{i, \droppedfeats})\}_{i=1}^{\numimp}) ] 
\end{align*}

Here, like in the original Shapley, each $\phi_j$ may be interpreted as the contribution of feature value $x_j$ to the shift $f(x)-\phi_0$. However in median-SHAP $\phi_0$ is the median prediction from $f$ in the reference population. Therefore comparing $f(x)$ with $\phi_0$ implies we're comparing our individual $x$ with the median \emph{predicted} individual i.e.\ the individual with median prediction in the cohort. Ultimately, with median-SHAP the anchor point is the median individual, and thus an observed data point. Using a conditional reference distribution further ensures that our black-box isn't evaluated off the data manifold i.e.\ concatenated instances aren't out of distribution. Contrastively, using a marginal reference distribution would jeopardise the interpretation by breaking the correlation and defining an out-of-distribution anchor point. Ultimately, using the median over references makes our estimation robust to outliers and skewed distribution compared to using the sample mean.

Given the clinical focus of our targeted application, we chose observational Shapley values to preserve the robustness to adversarial attacks and thus maintain the high safety standards needed for medical purposes. Further, note that in the case of a binary classification problems (e.g. having a survival time above/below a threshold) and assuming that the data set  $\{\blackbox(\localobs_{\includedfeats \cup j}, \imputed_{i, \droppedfeats/j})\}_{i=1}^{\numimp}$ is balanced, the median model outcome will be a realisation at the decision boundary. Such an approach is easier (but also more restricted) than a linear search for the decision boundary with e.g. $b$ perturbations used for LIME \cite{white2019measurable}. In contrast to $l1$ and $l2$ regularisation, additivity to $\blackbox(\localobs)$ of the Shapley values is preserved with median-SHAP. Moreover, while Shapley values lose their intuition when regularisation is added, the median Shapley values can be interpreted as the expected change in median model outcome when a feature is added.



\section{Experiments}
\label{sec:experiments}
We demonstrate the benefits of using median-SHAP by doing the following experiment on two datasets. We use the entire set as a training set, and a reference distribution.
\begin{enumerate}[leftmargin=4mm, nolistsep]
    \item Training a regression model $f$ on $(X,Y)$ where $X$ is a set of features and $Y$ is a median survival time. We denote $med = \text{Median}[f(X)]$.
    \item Using SHAP and median-SHAP to generate feature attributions for the outcome of a given individual $x$ .
    \item 'Re-labeling' the outcome of all individuals in the dataset to create a classification problem, such that: $Y'=1$ if $f(x)>med$ and $Y'=0$ otherwise.
    \item Training a classification model $g$ on $(X,Y')$.
    \item Using SHAP to compute feature attributions explanations for the classification of $x$ by our model $g$
\end{enumerate}
We repeat this experiment for randomly sampled individuals and report the relative differences (i) between median-SHAP for $f$ and SHAP for $g$ (ii) between SHAP for $f$ and SHAP for $g$. This process may attest if median-SHAP captures which are the most important features for predicting how an individual \emph{ranks} within the cohort, \wrt a model's predictions. In other words, if a feature is important for model $g$ it means that it is important to predict how our individual ranks \wrt the median individual. Therefore, if median-SHAP for $f$ computes feature attributions similar to the Shapley values for $g$, this confirms our method captures how the instance of interest compares with the anchor point, or median individual here. Details regarding the experiments and the datasets are shown in the Supplements \ref{sec:suppl_res}. 

\subsection{The Worcester Heart Attack Study}

The Worcester Heart Attack Study (Whas) is a longitudinal study on on acute myocardial infarction which looks at patient features at presentation \cite{goldberg2000decade}. A Random Survival Forest \cite{ishwaran2008random} -- which is a classification tree method for the analysis of right-censored survival data -- was trained on the entire set (N=500) to predict time of death, which occurred for 215 patients (43.0\%).

\begin{table}[H]
\centering
\resizebox{0.5\textwidth}{!}{%
\begin{tabular}{@{}|l|llll|@{}}
\toprule
Shapley type    & Age & BMI & Chf  & Heart Rate    \\ \midrule
\shortstack{SHAP on $f$ vs SHAP on $g$}           & 7.84  & 2.1  & -0.54 & 6.51  \\
\shortstack{median-SHAP on $f$ vs SHAP on $g$}    & 1.87  & -0.1 & -1.03 & 1.22  \\ \bottomrule
\end{tabular}%
}
\caption{Mean difference of scaled feature attributions for the Whas data comparing (i) mean Shapley values on $f$ vs mean Shapley values on the classification model $g$, (ii) median-SHAP on $f$  vs mean Shapley values on the classification model $g$.}
\label{tab:compar_whas}
\end{table}

Here, feature attributions should be interpreted comparing each instance of interest with the median patient, who is a 59 year old individual with a BMI of 29.3 $kg/m^2$, a heart rate of 117 bpm and who has experienced congestive heart failure. Table \ref{tab:compar_whas} shows the results of our experiment on the Whas dataset.

\subsection{The breast cancer survival data}
The breast cancer survival data \cite{desmedt2007strong} is a longitudinal study which explores the time dependence between a genetic prognostic signature and the occurrence of distant metastases. A Random Survival Forest was trained on 198 individuals to predict the time of potential metastases, which occurred for 51 patients (25.8\%). Resulting feature attributions are described in table \ref{tab:compar_breast}, and should be interpreted as comparisons with the median individual who has the following features: \verb|X202240 = 6.96|, \verb|X214806 = 7.32|, \verb|X219724_s = 6.33| and \verb|X203306_s = 10.51| and has a predicted median survival time of 51 time units. 

For both experiments, feature attributions computed using median-SHAP are closer to the ones for the classification task than the original mean Shapley values. This further demonstrates that median-SHAP is able to capture which features are predominant when predicting how our individual compares with the rest of the cohort, and more specifically here how it compares with the median individual. This further illustrates the applicability of our method in this context. Ultimately, note that the median is more appropriate as an estimator for such examples with a small sample size, due to its robustness to outliers.

\begin{table}[H]
\centering
\resizebox{0.5\textwidth}{!}{%
\begin{tabular}{@{}|l|llll|@{}}
\toprule
Shapley type    & X202240 & X219724\_s & X214806 & X203306\_s   \\ \midrule
\shortstack{SHAP on $f$ \\ vs SHAP on $g$}          & 0.04   & -3.32  & -2.54 & 10.51  \\
\shortstack{median-SHAP on $f$ \\ vs SHAP on $g$}     &  $\leq$ 0.001  & -1.41 & -3.57 & -1.28  \\ \bottomrule
\end{tabular}
}
\caption{Mean difference of scaled feature attributions for the breast cancer survival data comparing (i) mean Shapley values on $f$ vs mean Shapley values on the classification model $g$, (ii) median-SHAP on $f$  vs mean Shapley values on the classification model $g$.}
\label{tab:compar_breast}
\end{table}
\vspace*{-25pt}

\section{Concluding remarks}
\label{sec:discussion}

We introduce median-SHAP, a specific explanation method for survival models. In addition to the robustness of the median estimator \wrt outliers and its higher likelihood for skewed distributions, median-SHAP offers increased interpretability as comparisons are made with an actual individual from the reference population. Ultimately, we advocate that median-SHAP has improved clinical applicability as it allows physicians to illustrate their recommendations using person-centric explanations.


\newpage
\bibliography{bib}
\bibliographystyle{icml2022}

\newpage
\appendix
\onecolumn

\section*{Supplementary Material}

\section{Shaley axioms for qSHAP} 
\label{sec:suppl_axioms} 
Shapley values have been shown to satisfy the following axioms.
\begin{itemize}
\item According to the \emph{Dummy} axiom, a feature $j$ receives a zero attribution if it has no possible contribution, i.e. $\valuefct(S \cup j) = \valuefct(S)$ for all $S \subseteq \{1,...,\numfeats\}$. 
\item According to the \emph{Symmetry} axiom, two features that always have the same contribution receive equal attribution, i.e. $\valuefct(S \cup i) = \valuefct(S \cup j)$ for all $S$ not containing $i$ or $j$ then $\phi_i(v) = \phi_j(v)$. 
\item According to the \emph{Efficiency} axiom, the attributions of all features sum to the total value of all features. Formally, $\sum_j \phi_j(\valuefct) = \valuefct(\{1,..,\numfeats\})$. 
\item According to the \emph{Linearity} axiom, for any value function $v$ that is a linear combination of two other value functions $u$ and $w$ (i.e. $v(S) = \alpha u(S) + \beta w(S)$), the Shapley values of $v$ are equal to the corresponding linear combination of the Shapley values of $u$ and $w$ (i.e. $\phi_i(v) = \alpha \phi_i(u) + \beta \phi_i(w)$).
\end{itemize}
Since all these axioms have been defined conditionally on the value function $v$, they hold if the value function is changed. We have expressed qSHAP, and thus median-SHAP, as a change in the value function of standard Shapley values. As such they also adhere to the axioms. 

\section{Further experimental results} 
\label{sec:suppl_res} 

Black-box models were trained using the default parameters from \verb|sklearn| and \verb|scikit-survival| for the Multi-layer Perceptron regressor (MLP regressor) and the Random Survival Forest (RSF) respectively. For all experiments, we computed feature importances using the Mean Decrease Accuracy from the ELI5 framework (\url{https://github.com/mayur29/Machine-Learning-Model-Interpretation}. Only the top 4 features were kept and the model was retrained using this restricted set of features. 

\subsubsection{The Worcester Heart Attack dataset}
The dataset has 500 samples and 14 features. The endpoint is death, which occurred for 215 patients (43.0\%). A Random Survival Forest was trained on all samples. The model fit was measured with the out of bag score, which was equal to 0.761 for the entire model and 0.892 for the parsimonious model. The training set was used as a reference population for Shapley values computation. The URL for this dataset is: \url{https://scikit-survival.readthedocs.io/en/stable/api/generated/sksurv.datasets.load_whas500.html#sksurv.datasets.load_whas500}

\subsubsection{The breast cancer survival dataset}
The dataset has 198 samples and 80 features. The endpoint is the presence of distance metastases, which occurred for 51 patients (25.8\%). A Random Survival Forest was trained on all samples. The model fit was measured with the out of bag score, which was equal to 0.761 for the entire model and 0.844 for the parsimonious model. The training set was used as a reference population for Shapley values computation. The URL for this dataset is: \url{https://scikit-survival.readthedocs.io/en/latest/api/generated/sksurv.datasets.load_breast_cancer.html}

The median individual has \verb|X202240 = 6.96|, \verb|X214806 = 7.32|, \verb|X219724_s = 6.33| and \verb|X203306_s = 10.51| and has a predicted median survival time of 51 time units.

\end{document}